\documentclass{NLE}

\usepackage{amsmath}
\usepackage{graphicx}
\usepackage{natbib}
\ifpdf%
\usepackage{epstopdf}%
\else%
\fi
\usepackage{multirow}
\usepackage{times}
\usepackage{latexsym}
\usepackage{graphicx}
\usepackage{xcolor}
\usepackage{arabtex}
\usepackage{multirow}
\usepackage{makecell}
\usepackage{utf8}
\usepackage[utf8]{inputenc}
\setcode{utf8}
\usepackage[export]{adjustbox}

\usepackage[T1]{fontenc}
\usepackage[utf8]{inputenc}
\usepackage{microtype}
\usepackage{enumitem}
\usepackage{color, colortbl}
\definecolor{LightCyan}{rgb}{0.95,1,1}
\definecolor{LightGrey}{rgb}{0.90,0.90,0.90}

\begin{document}
\label{firstpage}

\lefttitle{\LaTeX\ Supplement}
\righttitle{Natural Language Engineering}

\papertitle{Article}

\jnlPage{1}{00}
\jnlDoiYr{2019}
\doival{10.1017/xxxxx}
\def\endabstract{\egroup}

\title{Emojis as Anchors to Detect \\Arabic Offensive Language and Hate Speech}

\begin{authgrp}
\author{Hamdy Mubarak\textsuperscript{1,*}, Sabit Hassan\textsuperscript{2}, and Shammur Absar Chowdhury\textsuperscript{1}}
\affiliation{\textsuperscript{1}Qatar Computing Research Institute, Hamad Bin Khalifa University, Doha, Qatar\\
\textsuperscript{2}School of Computing and Information,University of Pittsburgh, PA, USA\\
\textsuperscript{*}Corresponding Author. Email: \email{\textcolor{blue}{hmubarak@hbku.edu.qa}}}
        
\end{authgrp}

\history{(Received xx xxx xxx; revised xx xxx xxx; accepted xx xxx xxx)}

\begin{abstract}
We introduce a generic, language-independent method to collect a large percentage of offensive and hate tweets regardless of their topics or genres. We harness the extralinguistic information embedded in the emojis to collect a large number of offensive tweets. We apply the proposed method on Arabic tweets and compare it with English tweets -- analysing key cultural differences. We observed a constant usage of these emojis to represent offensiveness throughout different timespans on Twitter.
We manually annotate and publicly release the largest Arabic dataset for \textit{offensive}, \textit{ fine-grained hate speech}, \textit{vulgar} and \textit{violence} content. Furthermore, we benchmark the dataset for detecting offensiveness and hate speech using different transformer architectures and perform in-depth linguistic analysis. We evaluate our models on external datasets -- a Twitter dataset collected using a completely different method, and a multi-platform dataset containing comments from Twitter, YouTube and Facebook, for assessing generalization capability. Competitive results on these datasets suggest that the data collected using our method captures universal characteristics of offensive language.
Our findings also highlight the common words used in offensive communications, common targets for hate speech, specific patterns in violence tweets; and pinpoint common classification errors that can be attributed to limitations of NLP models. We observe that even state-of-the-art transformer models may fail to take into account culture, background and context or understand nuances present in real-world data such as sarcasm.

\end{abstract}

\maketitle

\section{Introduction}
\label{introduction}
{\small \textbf{Disclaimer}: \textit{Due to the nature of this work, some examples contain offensiveness, hate speech and profanity. This doesn't reflect authors' opinions by any mean. We hope this work can help in detecting and preventing spread of such harmful content.}}\\

Social media platforms provide a medium for individuals or a group to connect with the world and share their opinion (\cite{intapong2017assessing}), often with limited inhibitions. Taking the advantage of such liberation, social media users may use vulgar, pornographic, or hateful language. Such behaviour can result in the spread of verbal hostility and can impact users' psychological well-being (\cite{gulaccti2010effect,waldron2012harm}). \\

In recent years, Twitter has become highly popular in the Arab region (\cite{abdelali2020arabic}), with more than 27 million tweets per day (\cite{alshehri2018think}). Many look to Twitter to express their views, ideas and share their stories. While the importance of sharing these views and ideas is immense, the aforementioned societal problem of sharing malicious content also arise. \\

With the sheer volume of the content on social media, manually filtering out malicious content while maintaining users' right to the freedom of expression, is virtually impossible for the platform providers. The increased risk and effect of such hostility presence in social media has attracted many multidisciplinary researchers and motivated the need to automatically detect offensiveness of posts/comments and utilize the system to: \textit{(i)} filter adult content (\cite{cheng2014isc, mubarak2021adult}); \textit{(ii)} quantify the intensity of polarization (\cite{belcastro2020learning,conover2011political}); \textit{(iii)} classify trolls and propaganda accounts (\cite{alhazbi2020behavior, kareemsocinfo2017,dimitrov2021detecting,dimitrov2021task}); and \textit{(iv)} identify hate speech and conflicts (\cite{kiela2020hateful,davidson2017automated, ousidhoum2019multilingual,chung2019conan}). \\

Machine learning based automation approaches require labeled datasets to distinguish malicious content from others. Constructing datasets, however, is not a straightforward process. In a randomly sampled collection of tweets, the percentage of malicious tweets is very small. \citet{mubarak2017abusive} report that only 1-2\% of Arabic tweets are abusive. Percentage of hate speech is even smaller. This implies that to obtain a sizable dataset of malicious content, a massive number of tweets need to be labeled. To avoid this, existing approaches utilize heuristics to increase the percentage of malicious tweets before manually labeling them. Such heuristics include searching for specific language-dependent keywords or patterns. \newline

In this paper, we present an automated emoji-based approach of collecting tweets that have much higher percentage of malicious content, without having any language dependency. 
Emojis have quickly become an important part of our daily communication. They are used worldwide for conveying messages without any language or online platform barriers \cite{}; and has been referred to as an universal language (\cite{mei2019decoding, durscheid2017beyond}). Hence, the extralinguistic information carried by the emojis can be instrumental in capturing offensive content, without being impacted by the lack of language-dependent knowledge, or any preferred linguistic patterns. Using emojis also resolves challenges posed by non-standard spellings of offensive words. \\ 

To this end, we use emojis as anchors to collect tweets and manually annotate them for fine-grained abusive/offensive language categories -- including hate speech, vulgar, and violence tweet contents. We exploit the collected dataset for studying: (i) emoji usage in a different period of time; (ii) extraction of offensive words with different dialectal spellings and morphological variations; (iii) hate speech targets; (iv) linguistic content in vulgar and profanity words; and (v) exploring common patterns present in violent tweets.

Our comparative analysis shows that by using emojis as anchors, we end up with 35\% offensive (OFF) tweets and 11\% hate speech (HS) tweets, which is almost as double as the approach used by \citet{mubarak2017abusive}. Our experiments with various machine learning and deep learning classifiers show that the classifiers are able to learn effectively from the data. Further, we show that these classifiers are capable of generalizing well on two external datasets: \textit{i)} the OffensEval 2020 (SemEval, task 12) dataset (\cite{zampieri-etal-2020-semeval}) for Arabic offensive language identification, and \textit{ii)} the MPOLD dataset (\cite{chowdhury2020multi}) -- a multi-platform dataset containing news comments annotated for offensiveness from Twitter, YouTube and Facebook. This suggests our method captures universal characteristics of offensive language on social media and can be used to collect larger datasets with less effort and linguistic-knowledge support. \\

Therefore, the main contributions of the paper are:
\begin{itemize}
    \item We present an emoji-based method to collect offensive and hate speech tweets. We show that using the method, we can collect higher percentage of offensive and hate speech content compared to the existing methods. Moreover, we demonstrate that using this method we can collect large number of offensive contents in other languages such as Bengali.
    
    
    \item We manually label the largest dataset for offensiveness, along with fine-grained hate speech types, vulgarity (profanity) and violence. This labeled data will be publicly released for the community.\footnote{Data can be downloaded from this URL:  to be added} 
    
    \item We perform in-depth analysis of different properties of the dataset including common offensive words and hate speech targets. We show also that there are some culture differences in using emojis in Arabic offensive tweets compared to English.
    
    
    \item We build effective machine learning and deep learning based classifiers to automatically tag offensive and hate speech with high accuracy. We show that our model perform well on external datasets collected from Twitter and also demonstrate the potential generalisation capability of the designed model in different social media platforms -- such as YouTube and Facebook.
    \item We analyze common classification errors and provide some recommendations to enhance model explainability.
\end{itemize}

\noindent While our main focus, in this paper, is to create Arabic offensive dataset, this method can easily be adapted for other languages and potentially other text classification tasks such as sentiment analysis or emotion detection. \\


The paper is structured as follows. In section \ref{related}, we discuss related work, focusing on the methods used by other researchers to collect datasets for offensive language. In section \ref{data}, we describe our data collection and annotation jobs. Section \ref{analysis} contains extensive analysis of the dataset. The analysis includes study of offensive emojis, how emoji usage changes in a different time period. We further analyze offensive words and hate speech targets found in our dataset. In Section \ref{classifiers}, we train a set of machine learning and deep learning models for classification of offensive language and hate speech. We manually analyze errors made by our models and also use LIME explainability tool to interpret decisions made by our models. In this section we also show that models trained on our data achieves good results on other datasets. Section \ref{ethics} discusses ethics and social impact of our work. Finally, in Section \ref{conclusion}, we present summary of our findings.

\section{Related Work}
\label{related}



There has been a large amount of research in recent years to address and detect the growing use of `offensive languages' and hate speech in different social platforms (see \cite{nakov2021detecting} and \cite{salminen2020developing} for more details.). Detecting offensive language has been the focus of many shared tasks such as OffensEval 2020 (\cite{zampieri-etal-2020-semeval}) for five languages and OSACT 2020 (\cite{mubarak2020overview}) for Arabic. The best systems at the shared tasks (\cite{alami-etal-2020-lisac}, \citet{hassan-etal-2020-alt-semeval})  utilized Support Vector Machines and fine-tuned transformer models. Hate speech has been less explored compared to offensive language. A dataset consisting of 5\% hate speech was presented at OSACT  2020 shared task. The best system performed extensive preprocessing including normalizing \textbf{emojis} (translate their English
description to Arabic) and dialectal Arabic (DA) to Modern Standard Arabic (MSA) conversion among others (\cite{husain2020osact4}). ASAD (\cite{hassan-etal-2021-asad} is an online tool that utilizes the shared task datasets for offensiveness and hate speech detection in tweets along with other social media analysis components such as emotion (\cite{hassan2021crosslingual}) and spam detection (\cite{mubarak2020spam}).\\



\begin{table}
\caption{Available Arabic datasets along with the annotation labels, data source and collection method and percentage of offensive content. Sources: TW (Twitter), Aljazeera (AJ), FB (Facebook), YT (YouTube).}
\scriptsize
    \centering
    \begin{tabular}{l|l|l|r|l|l|l}
        \hline
        \rowcolor{LightGrey}
        \bf{Dataset} & \bf{Source} &  \bf{Method} & \bf{Size} & \bf{Language} & \bf{OFF\%} & \bf{Annotations (Values (1: Yes, 0: No))} \\\hline

        \cite{mubarak2017abusive} & TW & Controversial users & 1,100 & Egyptian & 59\% & OFF (1/0), Vulgar (1/0)\\
        & AJ & Deleted comments & 31,692 & MSA & 82\% & OFF (1/0), Vulgar (1/0)\\
        
        \rowcolor{LightCyan}
        \cite{albadi2018they} & TW & Keywords & 6,030 & MSA/DA & 45\% & Religious HS (1/0)\\
        
        \cite{alshaalan-al-khalifa-2020-hate} & TW & Keywords & 9,316 & Saudi & 28\% & HS (1/0)\\
        
        \rowcolor{LightCyan}
        \cite{mubarak2020overview} & TW & Pattern & 10,000 & MSA/DA & 19\% & OFF (1/0), HS (1/0)
        \\

        \cite{chowdhury2020multi} & TW, FB, YT & User replies to AJ & 4,000 & MSA/DA & 17\% & OFF (1/0), HS (1/0), Vulgar (1/0)\\ \hline
        
        \rowcolor{LightCyan}
        \bf{Our Method} & TW & Emojis & \bf{12,698} & MSA/DA & \bf{35\%} & \bf{OFF (1/0), Vulgar (1/0), Violence (1/0)}
        \\
        \rowcolor{LightCyan}
        & & & & & & \bf{HS (gender, race, ideology, social class,} \\
        \rowcolor{LightCyan}
        & & & & & & \bf{religion, disability)}\\
        
\hline
        \end{tabular}
    \label{tab:arabic-datasets}
    \vspace{0.5cm}
\end{table}

To collect potentially offensive tweets, some studies use a list of seed offensive words and hashtags (e.g. \citet{mubarak2017abusive}). This approach has several drawbacks: \textit{(i)} it is hard to maintain this list as offensive words are ever-evolving; \textit{(ii)} Arabic dialects are widely used on social media and building such a list for different dialects is a challenging task that requires deep knowledge about cultures in many countries;
\textit{(iii)} Arabic dialects have no standard writings. So, it is extremely difficult to list all possible surface forms and creative spellings for words in general including offensive words. Moreover, \textit{(iv)} Arabic has a rich morphology (both derivational and inflectional), and a large number of affixes can be attached to words -  e.g., \<وسيفعلها> (“wsyfElhA” – “and he will do it”)\footnote{We provide Arabic examples  and their Buckwalter transliteration and English translation throughout this paper.} -  which makes string match for offensive words less optimal. Finally, \textit{(v)} the offensiveness of many words is highly dependent on context, e.g., the word \<كلب> (“klb” – “dog”)  can be used in offensive contexts such as \<هو كلب> (“hw klb” – “He is a dog”), and in clean contexts such as \<عندي كلب> (“Endy klb” – “I have a dog”) . \\

Authors in \citet{mubarak2020arabic} showed that offensive language exists in less than 2\% of any random collection of tweets, and by considering a common pattern used in offensive communications, this ratio increases to 19\% (5\% are hate speech). This pattern is \<يا .. يا> (“yA .. yA” – “O .. O ..”) which is used mainly to direct the speech to a person or a group. This pattern is used across all dialects without any preference to topics or genres, however, it cannot be generalized to other languages. \\

In the same research, authors observed that the most frequent personal attack on Arabic Twitter is to call a person an animal name (i.e. direct name calling), and the most used animals are \<كلب> (“klb” – “dog”), \<حمار> (“HmAr” – “donkey”), and \<بهيم> (“bhym” – “beast”) among others. In \citet{chowdhury2020multi}, authors analyzed Arabic offensive language on Facebook, YouTube and Twitter, and they showed that some emojis are widely used in offensive communications including some animals (dog, pig, monkey, cow, etc.), some face emojis (anger, disgust, etc.) and others (shoe, etc.)\\

Unlike the aforementioned studies, we propose a generic method to collect offensive tweets regardless of their topics, genres or dialects. We applied it to Arabic Twitter and obtained the largest dataset of tweets labeled for offensiveness (approx. 13K tweets) with 35\% of the tweets labeled as offensive and 11\% as hate speech. In this method, we simply use a list of emojis that commonly appear in offensive communications including emojis that express anger, some animals and inanimate things extracted from existing datasets of offensive tweets. We believe with small modifications to this list, considering cultural differences, the approach can be used for other languages as well to collect a large number of offensive tweets. Comparison with available Arabic datasets for offensive language and hate speech is shown in Table \ref{tab:arabic-datasets}.\\
\begin{figure}
\begin{center}
\includegraphics{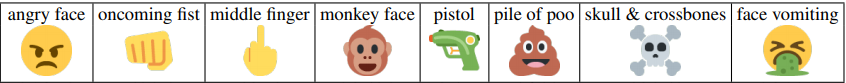} 
\vspace{0.5cm}
\caption{Emojis used in \cite{wiegand2021exploiting}}
\label{fig:emojis-angry-related}
\end{center}
\end{figure}

It is worth noting that authors in \citet{wiegand2021exploiting} used emojis to detect English abusive tweets. They used 8 emojis based on correlations between concepts and abusive language as reported in the literature. These emojis (Figure \ref{fig:emojis-angry-related}) indicate violence and death, anger and disgust, dehumanization, and disrespect. Authors mentioned that
middle finger is the strongest emoji as it is universally regarded as a deeply offensive gesture, and they used it to collect distantly-labeled training data. We will later highlight some differences between usage of offensive emojis in Arabic and English.

\section{Data}
\label{data}
\subsection{Data Collection}
\begin{figure}
\begin{center}
\scalebox{0.4}{
\includegraphics[]{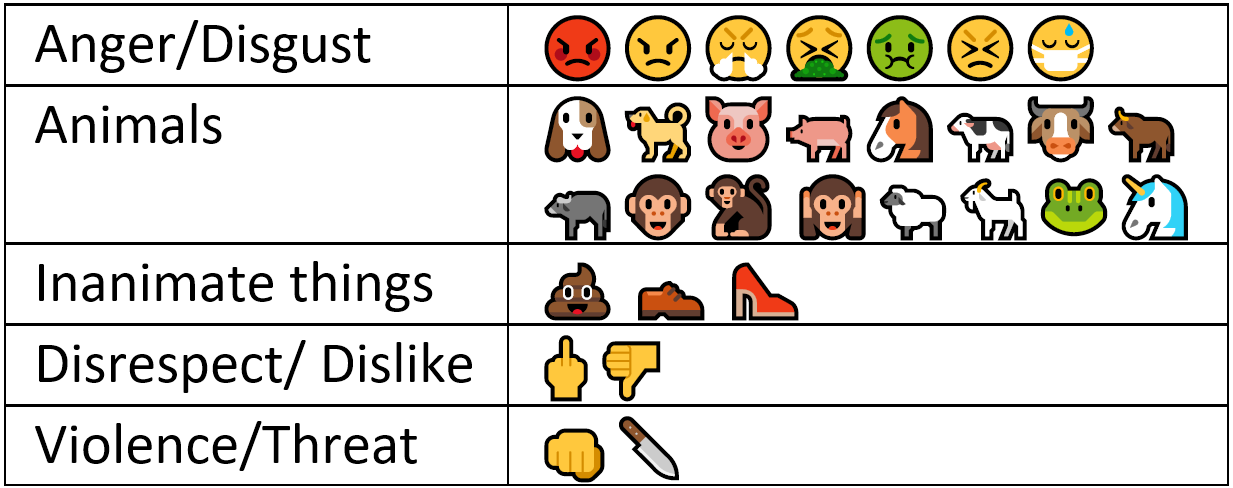} 
}
\vspace{0.5cm}
\end{center}
\caption{Categories of common offensive Emojis}
\label{fig:emojis-angry}
\end{figure}

We extracted common emojis that appear mostly in offensive communications from shared datasets in \citet{zampieri-etal-2020-semeval} and \citet{chowdhury2020multi}, and obtained their emoji variations from  https://emojipedia.org/. These emojis include some animals and symbols used for dehumanization and expressing disrespect, anger or disgust. 
The complete list of emojis used for the data collection and their categories is given in Figure \ref{fig:emojis-angry}. \\ 


From a collection of 4.4M Arabic tweets between June 2016 and November 2017, we extracted all tweets having any of the aforementioned emojis. After removing duplicates, near duplicates and very short tweets, we ended up with 12,698 tweets. 
We show later that tweets collected in another time period (in March 2021) still have high percentages of offensiveness and hate speech.

\subsection{Annotation and Quality Control}
\label{off_annotation}
We created an annotation job on Appen crowdsourcing platform to judge whether a tweet is offensive or not (\textit{Job1}) and we invited annotators from all Arab countries.\footnote{We described offensive language to annotators as any rude and socially unaccepted language used by some users to offend a person or a group. This  includes also vulgar and swear words.} Quality was assured using 200 questions that we manually obtained gold labels for (hidden test questions). Annotators should pass 80\% of them to continue.\footnote{We paid \$15 per hour of work to conform to the minimum wage rate in the US.} Each tweet was judged by 3 annotators and more than 190 annotators contributed to this job. Such a large number is needed for a subjective task like judging tweet offensiveness. Inter-Annotator Agreement (IAA) using  Cohen’s kappa ($\kappa$) value was 0.82, which indicates high quality annotations (\cite{LandisKoch_Kappa10.2307/2529310}).\\

Annotators fully agree (3 out of 3) in 65\% of the tweets. For better quality, we hired and trained an expert linguist who is familiar with different dialects to carefully check all tweets having different opinions. This resulted in changing 18\% of the labels (w.r.t to the previously assigned majority label) for the tweets. 

\subsection{Annotation of Hate Speech, Profanity and Violence}
We created another job on Appen with all the annotated offensive tweets from \textit{Job1} and asked annotators to detect the presence of hate speech. We defined hate speech as any content that contains offensive language targeting group of people based on common characteristics such as: race/ethnicity/nationality, religion/belief, ideology, disability/disease, social class, and gender. We used job settings similar to Job1. 
Additionally, we asked annotators to judge whether the  offensive tweets have profanity or vulgar words, and whether they promote violence. Basic statistics and examples from different annotations are listed in Table \ref{tab:stats}. 

\begin{table*}[!htb]
\caption{Statistics and examples from the annotated corpus (Total of 12,698 tweets)}
\small
    \centering
    \scalebox{0.93}{
    \begin{tabular}{|l|r|r|r|}
        \hline
        \rowcolor{LightGrey}
        \multicolumn{1}{c|}{\bf{Class/Subclass}} & \multicolumn{1}{c|}{\bf{\#}} & \multicolumn{1}{c|}{\bf{\%}} & \multicolumn{1}{c|}{\bf{Example}}\\\hline
        \bf{Clean} & 8,235 & 65 & \<لن تحــصــل علــى غــدٍ افــضل مادمــت تفــكر بالامــس> \\
        (CLN)& & & You won't have a better tomorrow as long as you think about yesterday.\\\hline \hline
        \bf{Offensive} & 4,463 & 35 & \<يلعن ابوك على هالسؤال. عساه ينقرض الكريه >\\
        (OFF)& & & May God curse your father for this question! I hope this fool will die out!
        \\\hline
        \rowcolor{LightCyan}
        * Hate Speech & 1,339 & 11 & (Note: 30\% of Offensive tweets are labeled as Hate Speech)\\\hline
        - Gender  & 641 & 48 &  \<بنات اليوم قليلات أدب. والله ما نوصل لعهر بعض الرجال> \\
        & & & Girls today are impolite. I swear to God, we don't reach for some men immorality. \\\hline
        - Race  & 366 & 27 & \<شعبكم متخلف. الله ياخذك إنتي والفلبين>\\
        & & & People of your country are musty. May God take (kill) you and the Philippines.\\ \hline
        - Ideology  & 190 & 14 & \<ناديك وضيع لا شك في ذلك. حزبك لا يقدر إلا على النباح>  \\
        & & & You club is vile, no doubt about that. Your party cannot do anything except barking. \\\hline        
        - Social Class  & 101 & 8 & \<دامك مقيم انكتم وخل اهل الأرض الاصليين يتكلمون. ابلع يا سباك!>  \\
        & & &  As you are a resident, shut up and let original citizens speak.  Swallow, plumber!\\
        \hline        
        - Religion  & 38 & 3 & \<إنتوا بتعملوا ف ديك أبونا كده ليه هو إحنا كفرة ولايهود>  \\
        & & & Why are you doing this to us? Are we disbelievers or Jews? \\\hline        
        - Disability  & 3 & 0 & \<ذا القزم طلعت جايزتن له بس ماعرف يعبر>  \\
        & & & This dwarf got two prizes, but he does not know how to express.\\\hline

        \rowcolor{LightCyan}
        * Vulgar & 189 & 1.5 & \<كاتبة ع حسابك قحبة بنت قحبة و بتتعجبي انهم بيبعثو صور فاضحة>  \\
        \rowcolor{LightCyan}
        & & & \<* قحبة: امرأة فاجرة فاسِدة تمارس البِغاء (معجم اللغة العربية المعاصر)>\\
        \rowcolor{LightCyan}
        & & & You describe yourself as a prostitute, and you wonder why they're sending porn photos!\\\hline

        \rowcolor{LightCyan}
        * Violence & 85 & 0.7 & \<صفعه بهذا الحذاء حجم راسك على قرعتك. ييييععع ودي اطعنها>  \\
        \rowcolor{LightCyan}
        & & & Slap by this shoe on your bald head. Yucky! I'd like to stab her.\\\hline
        \rowcolor{LightCyan}
        \end{tabular}}
    \label{tab:stats}
\end{table*}

\section{Analysis}
\label{analysis}
We obtained a total of 582 different emojis in the final annotated data. Out of those, 122 emojis appear more than 20 times. In this section, we first report some observations seen in our data, followed by in-depth analysis of emoji in a different time period, linguistic characteristics associated with emojis along with targets of hate speech in these tweets.

\subsection{Common Offensive Emojis}
\begin{figure}
\begin{center}
\includegraphics[scale=0.80]{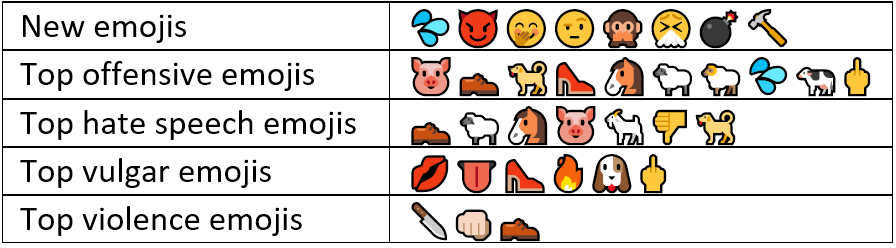} 
\vspace{0.7cm}
\caption{Categories of common offensive Emojis. Emojis in the same row are sorted based on percentage of offensive tweets (in a descending order).}
\label{fig:emojis-angry2}
\end{center}
\end{figure}

We report our observations about common emojis used to represent offensiveness in below:

\begin{itemize}[leftmargin=*]
  \item There are additional emojis, unique to our initial seed list, frequently used in the annotated offensive tweets. This class of emojis (see ``New emojis'' row in Figure \ref{fig:emojis-angry2}) includes instances like spit/drops (65\% of tweets having it was labeled as offensive) to hammer (18\% offensive). 
  This suggests a potential for enriching the initial emoji seed list and expanding the dataset in an iterative way.
  

\item Top offensive emojis are shown in Figure \ref{fig:emojis-angry2} (mostly animals). If a tweet has any of them, most likely it is labeled as offensive (pig (84\% offensive) to middle finger (62\%)). Similar emojis appear in hate speech tweets with percentages range from 36\% to 18\% respectively.

\item Top vulgar emojis are mostly used in tweets having adult content (e.g. adult ads).

\item For violence category, the most common emojis are: knife, punch and shoe in order.
\end{itemize}

\subsection{Emojis Usage In A Different Time Period}
We collected tweets from a different time period (in March 2021) to study the usage pattern of offensive emojis over time. For this, we selected six emojis from Figure \ref{fig:emojis-angry2} and their corresponding words to extract their tweets.\footnote{We ignored "high heel shoe" emoji as it is used dominantly in adult ads, and "donkey/horse" emoji as it is ambiguous in its usage (as donkey in some cases and horse in others). ``Donkey'' emoji is not found in https://emojipedia.org} For each emoji and its corresponding word, we then randomly selected 200 tweets and asked an Arabic native speaker to judge the content for offensiveness.\footnote{Some words appear less than 200 times in this collection.}\\

\begin{figure}[!ht]
\begin{center}
\includegraphics[scale=0.80]{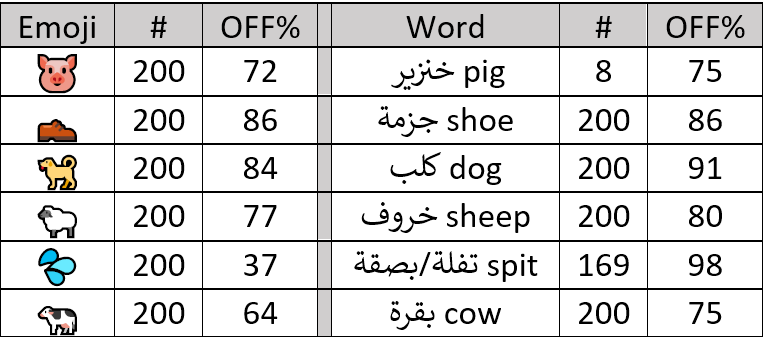} 
\vspace{0.5cm}
\caption{Percentage of offensive tweets for emojis and their corresponding words in samples from March 2021}
\label{fig:emojis-angry3}
\end{center}
\end{figure}

\begin{figure}[!h]
\begin{center}
\includegraphics[scale=0.7, frame]{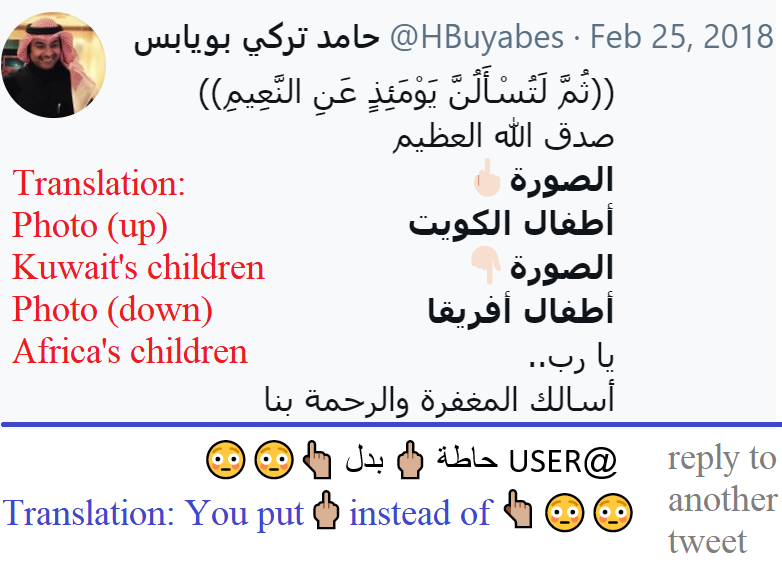}
\vspace{0.5cm}
\caption{Wrong usage of middle finger emoji in Arabic tweets}
\label{fig:emojis-angry-middle-finger}
\end{center}
\end{figure}

Our study suggests that  some emojis (see Figure \ref{fig:emojis-angry3}) and their corresponding words are widely used in offensive tweets during different periods of time.
While middle finger was reported as the strongest offensive emoji in English tweets (\cite{wiegand2021exploiting}), we found that there are more common offensive emojis appear in our dataset of Arabic tweets such as pig (84\%), shoe (77\%) and dog (68\%) (first in ``Top offensive emojis'' row in Figure \ref{fig:emojis-angry3}). Tweets having middle finger emoji were tagged as offensive in 62\% of the cases. We found annotation errors due to short context in some cases, but some users use middle finger mistakenly instead of index finger\footnote{We anticipate middle finger emoji is not widely understood as offensive by all Arab users.} as shown in Figure \ref{fig:emojis-angry-middle-finger}.

\subsection{Emojis and Linguistic Usage}

\begin{figure}
\begin{center}
\includegraphics[scale=0.75]{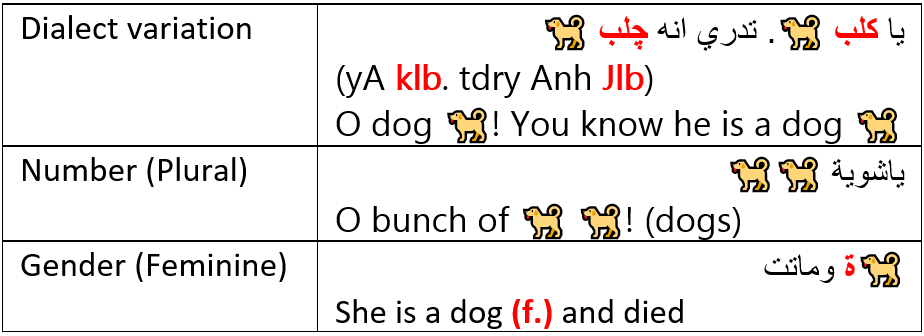}
\vspace{0.5cm}
\caption{Emojis co-occur with dialectal and morphological variations}
\label{fig:emojis-angry-variation}
\end{center}
\end{figure}

As observed in \cite{donato2017investigating}, emojis are often redundant and convey something already expressed verbally in tweets. Thus, offensive emojis may co-occur with many offensive words. As dialectal Arabic (DA) is widely used on Twitter, it's very hard to list all variations of offensive dialectal words. Figure \ref{fig:emojis-angry-variation} shows an example where offensive emojis co-occurred with offensive dialectal words which confirms the aforementioned observation. It also shows how emojis are used to cover morphological variations such as number and gender.

\subsection{Offensive Words}
To partially solve spelling mistakes in the collected tweets, we normalized some letters which are commonly used interchangeably by mistake, namely letters \<أإآ،ة،ى> to letters  \<ا،ه،ي> in order. We calculated valence score for normalized words in offensive and clean tweets as described in \citet{conover2011political}. The score helps determine the distinctiveness of a given word in a specific class in reference to other classes. Given $N(t,off)$ and $N(t,cln)$, which are the frequency of the term $t$ in offensive tweets ($off$) and clean tweets ($cln$) in order, valence score is computed as: 


\begin{equation}
\large
V(t) = 2* \frac{\frac{N(t, off)}{N(off)}}{\frac{N(t, off)}{N(off)} + \frac{N(t, cln)}{N(cln)}} - 1
\end{equation}

\noindent  where $N(off)$ and $N(cln)$ are the total number of occurrences of all words in offensive ($off$) and clean ($cln$) tweets respectively.\\

    

\begin{figure}
\begin{center}
\includegraphics[scale=0.40]{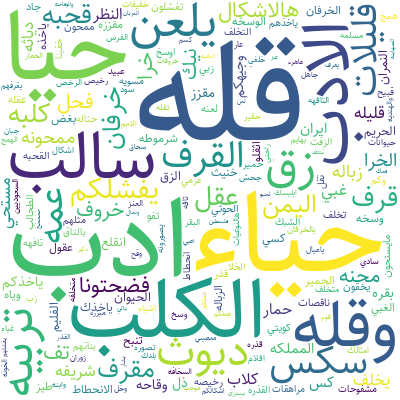} 
\vspace{0.3cm}
\caption{Offensive words (including vulgar words)}
\label{fig:emojis-OFF-words}
\end{center}
\end{figure}

Figure \ref{fig:emojis-OFF-words} shows top words with the highest valance score in offensive tweets.\footnote{Condition: Valence score >= 0.8 and frequency >= 5.} In addition to some animal words (e.g. dog, sheep), the list contains many dialectal words that are widely used in offensive communications such as \<زق، خرا، زباله> (``zq, xrA, zbAlh'' - shit, garbage) and some adult ads words such as \<سكس، ديوث> (``sks, dywv'' - sex, cuckold). This proves the efficiency of starting with offensive emojis to collect related offensive words without any preference to dialect or genre. All lists of extracted words with their valence scores will be made publicly available.\\

For intrinsic evaluation, we manually revised these top offensive words, and we found that 71\% of them appear dominantly in offensive communications and can be considered as correct offensive words. However, the remaining 29\% are not necessarily offensive and can appear in many clean contexts. This includes named-entities such as \<إيران، بوتين، الحوثي، بشار> (``AyrAn, bwtyn, AlHwvy, b\$Ar'' - Iran, Putin, Houthi, Bashar) in addition to some words that were concentrated  by chance in our offensive tweets. Examples of such words are: \<ذل، بلدك، ذكور> (``*l, bldk, *kwr'' - humiliation, your country, males).

\subsection{Hate Speech Targets}
We used the aforementioned valence score formula to extract common words appear in all hate speech tweets. Then, we manually filtered and grouped them to extract common targets for hate speech. These targets include women, countries, groups and political parties as listed in Table \ref{tab:hate-speech-targets}.

\begin{table*}[!htb]
    \caption{Common targets in hate speech tweets}

    \centering
    \scalebox{0.7}{
    \begin{tabular}{|l|r|r|}
        \hline
        \rowcolor{LightGrey}
        \multicolumn{1}{c|}{\bf{Target}} & \multicolumn{1}{c|}{\bf{Words}} &  \multicolumn{1}{c|}{\bf{Example}}\\\hline
        Women & \<الحريم، البنات>  &\<الله يسامح اللي سمح للحريم يدخلون الملعب. منظر مقرف!> \\
        & women, girls & May God forgive those who allowed women to enter the stadium. Disgust!\\\hline

        \rowcolor{LightCyan}
        Iran & \<إيران، الفرس، المجوس> & \<كل زق أنت وإيران المجوسية>\\
        \rowcolor{LightCyan}
        & Iran, Persians, Magi & Eat shit, you and the Magian  Iran.\\\hline
        
        Israel & \<إسرائيل، الصهاينة>& \<اللعنة عليك وعلى إسرائيل>\\
        & Israel, Zionists & May God curse you and Israel! \\\hline
        
        \rowcolor{LightCyan}
        Jews & \<اليهود> & \<أنتم تقتلون الأبرياء وتتركون اليهود يا خونة> \\
        \rowcolor{LightCyan}
        & Jews & You kill innocent people and leave Jews. O traitors!\\\hline
        
        Houthi & \<الحوثي، الحوثيين> & \<دمرتم كل جميل في وطني. الحوثي عدو الانسانية> \\
        (group) & Houthi, Houthis & You destroyed all the beauties in my country. Houthi is the enemy of humanity. \\\hline
        
        \rowcolor{LightCyan}
        Saudi Arabia & \<السعودية، السعوديين> & \<السعودية من الاحتضار الديني الى حضارة العري>\\
        \rowcolor{LightCyan}
        & KSA, Saudis & Saudi Arabia from religious agony to a civilization of nudity\\\hline
        
        Muslim Brotherhood & \<الإخوان>& \<هادو المرتزقة من أتباع حماس والإخوان> \\
        (group, party)& Brotherhood & These mercenaries are followers of Hamas and the Brotherhood \\\hline
        
        \rowcolor{LightCyan}
        Qatar & \<قطر، القطريين> & \<اعلام دويلة قطر تنبح ليل نهار> \\
        \rowcolor{LightCyan}
        & Qatar, Qataris & Media in the tiny country of Qatar barks day and night \\\hline

        USA & \<أمريكا، الأمريكان> & \<خرا عليك وعلى روسيا وأمريكا وخود الخليج بمعيتك> \\
        & America, Americans& Shit on you and on Russia and the US in addition to the Gulf. \\\hline

        \rowcolor{LightCyan}
        Turkey & \<تركيا، الأتراك، الترك، العثمانيين> & \<أنت يا تركي كذاب مثل أجدادك>\\
        \rowcolor{LightCyan}
        & Turkey, Turks, Ottomans & You, Turkish, are a liar like your ancestors. \\\hline
        
        \end{tabular}}
    \label{tab:hate-speech-targets}
\end{table*}

\subsubsection{Religious Hate Speech Targets}
\begin{figure}
\begin{center}
\includegraphics[scale=0.45]{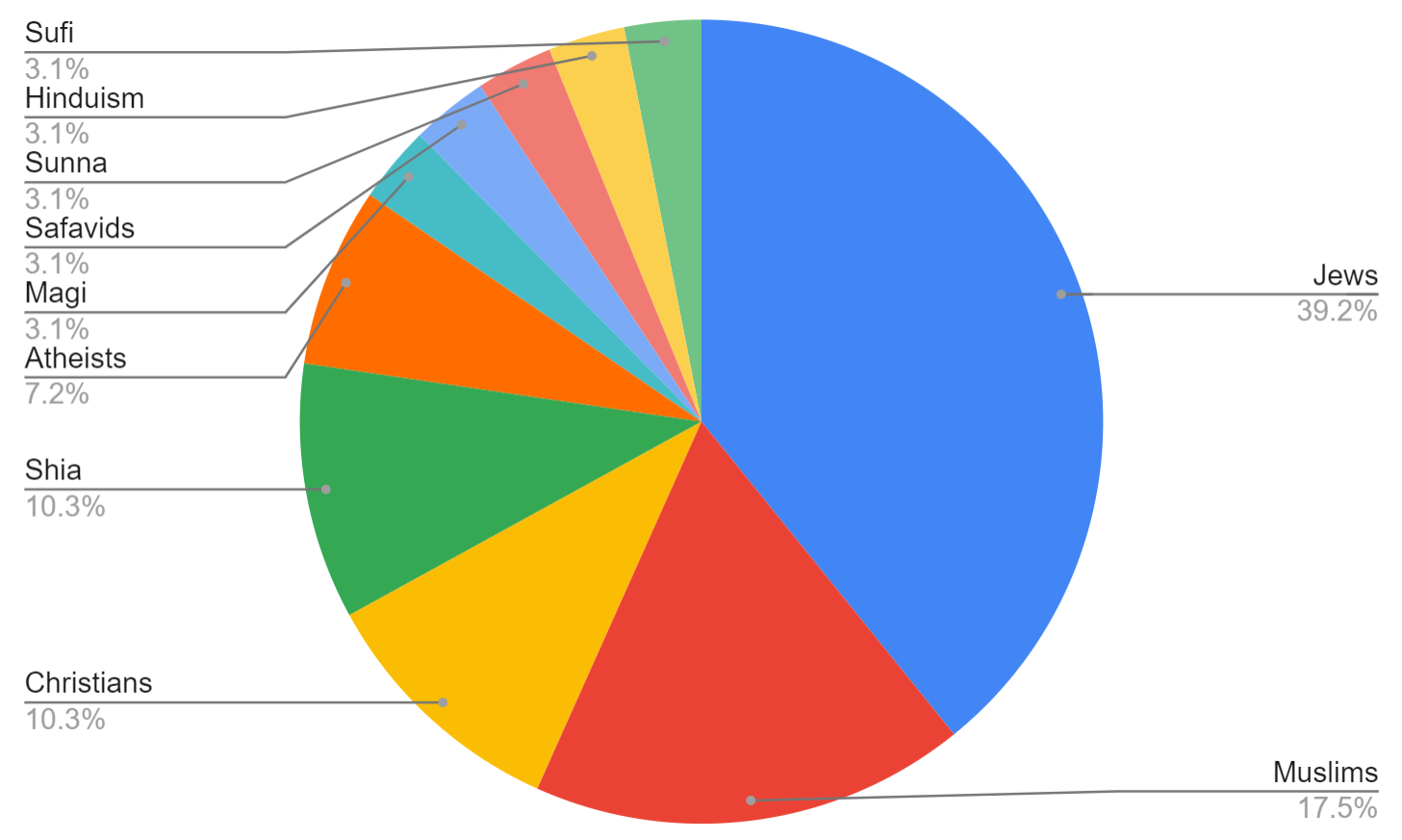}
\vspace{0.3cm}
\caption{Hate speech\% for different religious groups}
\label{fig:emojis-angry-hate-speech.png}
\end{center}
\end{figure}

Detecting religious hate speech in Arabic was the main focus of \citet{albadi2018they}. Authors mentioned that Arabic is the official language in six of the eleven countries with the highest Social Hostilities Index.\footnote{SHI is measure of crimes partially motivated
by religion.} They found that 33\% of all hateful tweets targeted against Jews followed by Shia and Christians. We analyzed hateful tweets in our dataset and we found similar results where Jews are the main target of religious prejudice with 39\% of all hateful tweets followed by Muslims, Christians and Shia. Figure  \ref{fig:emojis-angry-hate-speech.png} shows distribution of hate speech tweets for different religious groups.

\subsection{Vulgar and Profanity Words}
We observe that vulgar words include genitals, sexual actions and references in addition to words are used in adult ads. They are mainly written in DA and the most common words are shown in Figure \ref{fig:emojis-OFF-words}.

\subsection{Violence Tweets}
As violence tweets are rare (0.7\% in our collection), we manually analyzed them all to extract common patterns. These patterns are summarized in Table \ref{tab:violence-patterns}. 
First, we define \textit{<head>} for head and neck, and \textit{<body>} for body parts including \textit{<head>} that can be used as objects for many violence verbs, ex: I will stab [him/his neck]. For future work, we plan to extend this list using synonyms and related words from dictionaries/BERT models. Then, we extract tweets having any of these verbs and targets after applying some morphological expansions.

\begin{table}[!htb]
\caption{Common patterns in violence tweets}
    \centering
    \begin{tabular}{|l|r|}
        \hline
        \rowcolor{LightGrey}
        \bf{Pattern. S:Sbj, V:Verb, O:Obj} & \multicolumn{1}{c|}{\textbf{Description/Verbs}} \\\hline
        
        \rowcolor{LightCyan}
        \textit{<head>} & \<رأس، دماغ، وجه، رقبة> \\
        \rowcolor{LightCyan}
        & head, brain, face, neck\\
        \hline
        
        \rowcolor{LightCyan}
        \textit{<body>} & \<رأس، دماغ، وجه، رقبة، عين، خشم، أسنان، بطن>\\
        \rowcolor{LightCyan}
        & head, brain, face, neck, eye, nose, teeth, stomach
        \\\hline
        
        S V O, O: \textit{<human>} & \<يقتل، يذبح، يدبح>\\
        (ex: I will kill you) & kill, slaughter\\\hline
        S V O, O: \textit{<human>/<body>} & \<يدوس، يدعس، يجلد، يطعن، يضرب، يكسر>\\ 
        (ex: I will hit your head) & trample, whip, stab, hit, break\\
        \hline
        S V O, O: \textit{<head>} & \<يقطع، يفتح، يطير>\\
        (ex: I will cut your neck) & cut\\\hline
        
        \textit{<hit>} on \textit{<body>} & \<كف، جزمة، صفعة: على>\\
        (ex: A slap on your face) & shoe, slap\\\hline

        \end{tabular}
    \label{tab:violence-patterns}
\end{table}

\subsection{Is Emojis A Universal Anchor?}

Typical approach to crawl data from social media, specially from Twitter, is heavily dependent upon task-specific keywords. Preparing such keyword-specific anchors needs a significant knowledge on the language usage. However, using emojis to collect the data removes the initial language dependency and inherited bias from the keyword-search space.

In this Section, we study the efficacy of emojis to collect offensive tweets for other languages. We utilize the proposed pipelines for Bengali (a low-resource language)\footnote{Bengali is the lingua franca of Bengalidesh and parts of India. There are 228 million native speakers, making it the fifth most-spoken native language in the world. Source: https://en.wikipedia.org/wiki/$Bengali_language$}  and analysed the percentage of offensive tweets obtained using the method.
Moreover, we also include analysis on English tweets to indicate how offensive communication and emoji usage are still culture-dependent, if not language.



\begin{figure}[!htb]
\begin{center}
\includegraphics[scale=0.5, frame]{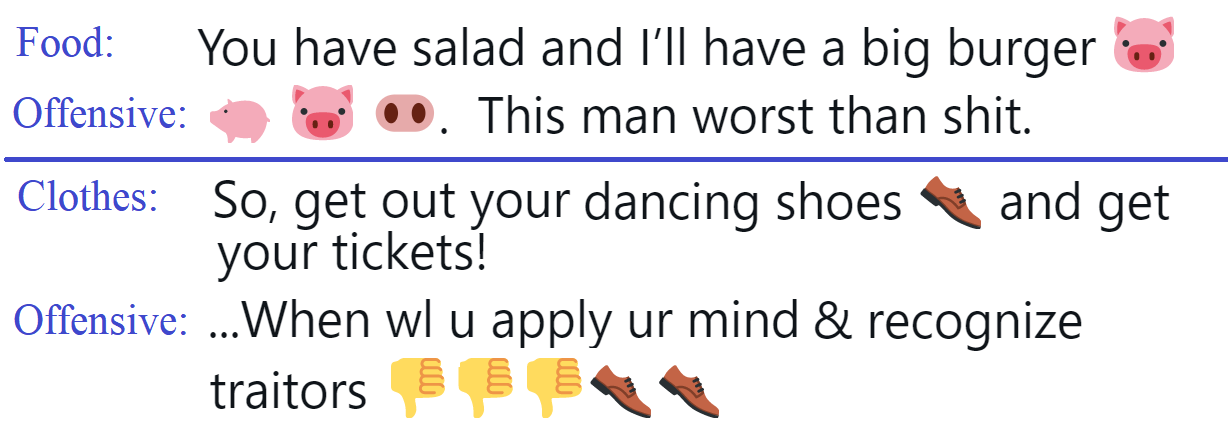}

\vspace{0.5cm}
\caption{English examples of pig and shoe emojis (topics of common usages and offensive contexts are shown)}
\label{fig:en-pig-shoe}
\end{center}
\end{figure}

\begin{figure}[!htb]
\begin{center}
\scalebox{0.55}{
\includegraphics[]{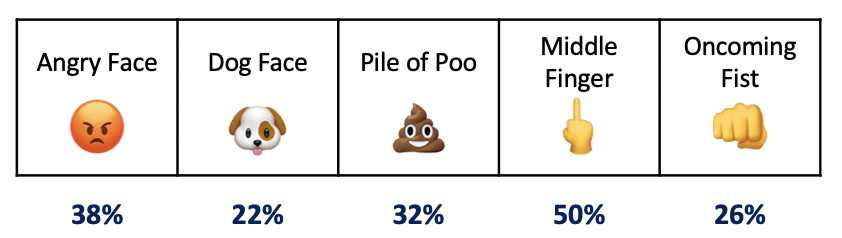}}
\vspace{0.6cm}
\caption{Total percentage of offensive tweets in a sample of 50 tweets for each emoji in Bengali tweets.}
\label{fig:bn-pig-shoe}
\end{center}
\end{figure}

\begin{figure}[!htb]
\begin{center}
\scalebox{0.45}{
\includegraphics[frame]{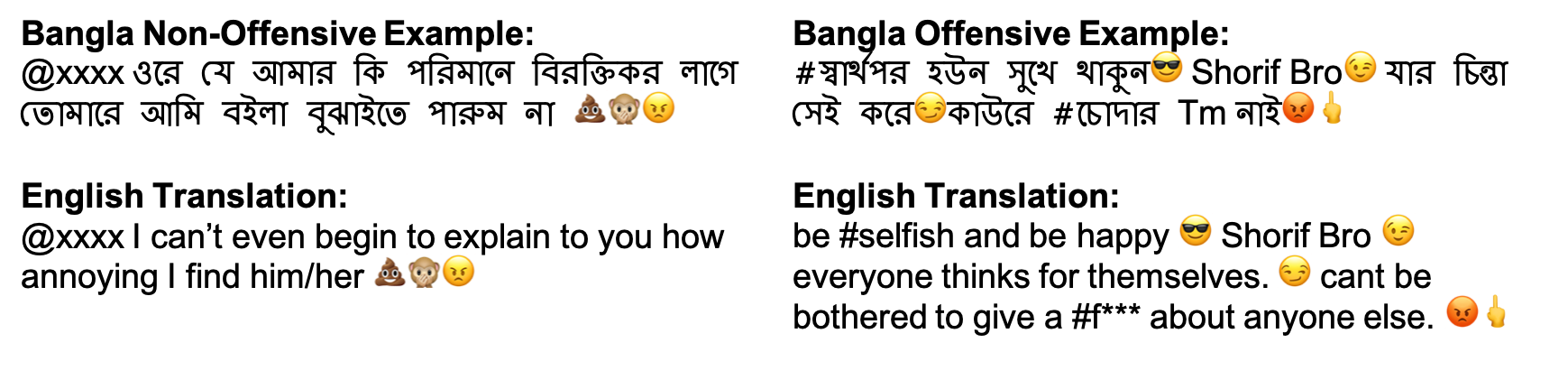}}
\vspace{0.6cm}
\caption{Examples of (non-)offensive tweets in Bengali collected using emojis.}
\label{fig:bn-example}
\end{center}
\end{figure}

For Bengali, we collected 200 tweets in August 2021 for five emojis (a total of 1K tweets) from the different groups listed in Figure \ref{fig:emojis-angry}. From these 200, we randomly selected 50 tweets per emoji (a total of 250 tweets)  
and a Bengali native speaker annotated them for offensiveness. We noticed that $\approx 34\%$ of all the annotated tweets are offensive. We observed that tweets with `middle finger' had most frequent ($50\%$ of tweets) number of offensive content, followed by `pile of poo' emoji (32\% of tweets). The detailed distribution is in Figure \ref{fig:bn-pig-shoe} with examples in Figure \ref{fig:bn-example}. We plan to use this method to collect and annotate offensive tweets for other low-resource languages such as Hindi,\footnote{Pilot annotation of Hindi tweets shows similar trends (52\% offensive for `middle finger', and 40\% offensive for `angry face'.)} in addition to Bengali. We keep this as a potential future work.\\


As usage of emojis for offensive content in English is well-studied, we only collected English tweets having the top Arabic offensive emojis (pig and shoe) in April 2021. We collected more than 8,000 tweets per each, and we randomly selected and annotated 200 tweets for each emoji. We noticed that less than 5\% of those tweets are offensive. Often, pig emoji appears in contexts about food while shoe emoji appears in tweets talking about fashion and clothes. Examples are shown in Figure \ref{fig:en-pig-shoe}. \\

This suggests that while some emojis are widely used in offensive communications in a particular culture, they might be used completely in different ways in other languages or cultures. Thus, it's important to keep these differences in mind when collecting data using emojis from different cultures.

\section{Experiments and Results}
\label{classifiers}

In this Section, we describe the details of machine learning algorithms - SVM and transformers, used to test the performance of the dataset followed by the experimental setup. We then present out classification results for in-domain model testing along in-detailed classification error analysis and model prediction explanability.
Furthermore, we evaluated the model's generalization capability across different datasets and various social media platforms.


\begin{table}[!htb]
    \caption{Distribution of offensive and hate speech data}
    \centering
    \scalebox{0.9}{
    \begin{tabular}{|c||c|c|c||c|}
    \hline
    & \bf{Train} & \bf{Dev} & \bf{Test} & \bf{Total}
 \\ \hline\hline
    \textit{OFF}	& 3,172 & 404 & 887 & 4,463\\
    \hline
    \textit{NOT\_OFF} &	5,716 &	865 &	1,654 &	8,235\\
    \hline \hline
    \textit{HS} & 	959 &	109 & 	271 &	1,339\\
    \hline
    \textit{NOT\_HS} &	7,929 & 1,160	& 2,270	& 11,359\\
    \hline\hline
    \rowcolor{LightCyan}
    \bf{Total}	& 8,888	& 1,269	& 2,541 & 12,698\\
    \hline
    \end{tabular}
    }
    \label{tab:distribution}
\end{table}

\subsection{Classification Models}
\label{ssec:architectures}
In order to design the automated classification system, we fine-tuned the state-of-the-art monolingual (Arabic) pretrained transformer architectures -- AraBERT (\cite{Antoun2020AraBERT}) and QARiB (\cite{abdelali2021pretraining}). Furthermore, we compared the performance of the monolingual transformers with two multilingual BERT: mBERT (\cite{devlin-2019-bert}) and XLM-RoBERTa (\cite{conneau2019unsupervised}). 
As a baseline, we use a SVM (\cite{platt1998sequential}) classifier, due to its reputation as an universal learner and for its ability to learn independently of the dimensionality of the feature space and the class distribution of the training dataset.\\


        

\paragraph{Support Vector Machines (SVM)} We used word n-gram and character n-gram vectors weighted by term frequency-inverse document frequency (tf-idf) as features to SVM. We experimented with character n-grams ($n=[2,5]$) and word n-gram ($n=[1,3]$) individually and also combination of them using the linear kernal with default parameters of SVM classifier ($C=1.0$).\\

\paragraph{Transformer models}
We fine-tuned the mono- and multilingual transformer models on our training data. 
For the monolingual task, both the transformers -- AraBERT and QARiB -- are of identical architecture with variation in the pretraining data. AraBERT is trained on Arabic Wikipedia, whereas QARiB is trained on Twitter data and Arabic Gigaword, thus encapsulating formal and informal writing styles in each transformers.\\

As for the multilingual models, we fine-tuned mBERT, trained on Wikipedia articles for 104 languages including Arabic using the case sensitive base model. In addition, we also fine-tuned XLM-RoBERTa -- trained on a large dataset with 100 languages using cleaned CommonCrawl data.\\ 

\subsection{Experimental Data and Setup}
\paragraph{Data Split} We trained the classifiers using 70\% of the data, and validated with 10\%.  
We used the rest (20\% of the data) for testing the system performance. Detailed data distribution of offensive and hate speech tweets are given in Table \ref {tab:distribution}.

\paragraph{Evaluation Measures} We evaluated the dataset using macro-averaged precision, recall and F1-measure, in addition to the accuracy. 

\paragraph{Offensive/Hate Speech Classifiers} For the respective downstream tasks, we fine-tuned the aforementioned BERT models by adding a dense layer that outputs class probabilities. We use a learning rate of $8e-5$ with a batch size of $64$ and $3$ epochs. For the fine-tuning, we restricted the  maximum input length to 47 tokens ($99^{th}$ percentile in training dataset), with no extra preprocessing of the data.

\subsection{Results}
\label{ssec:result}

From the reported results, presented in Table \ref{tab:offensive} and Table \ref{tab:hate}, we observe the monolingual models significantly outperforms the multilingual models. This observation is in-aligned with previous studies (\citet{polignano2019alberto,chowdhury2020improving}). 

\begin{table}[]
    \caption{Macro-averaged (P)recision, (R)ecall and F1 for \textbf{Offensive language} classification. * represent results obtained with different learning rate (2e-5) instead of 8e-5. C: Character n-grams W: Words n-grams.}
    \centering
    \scalebox{0.9}{
    \begin{tabular}{|c|c|c|c|c|}
        \hline

        \bf{Classifier} & \bf{Acc\%} & \bf{P} & \bf{R} & \bf{F1} \\ \hline\hline
        SVM (C) & 78.32 & 76.75 & 74.06 & 74.99\\
        \hline
        SVM (W) & 72.14 & 69.82 & 70.81 & 70.15\\
        \hline
        SVM (C+W) & 78.16 & 76.18 & 74.75 & 75.33\\
        \hline\hline
        mBERT & 76.43 & 74.09 & 73.32 & 73.66 \\
        \hline
        XLM-RoBERTa$^*$ & 75.00 & 72.50 & 72.47 & 72.48\\
        
        \hline
        \hline
        AraBERT & 82.09 & 80.50 & 79.63 & 80.02\\
        \hline
        QARiB  & \bf{84.02} & \bf{82.53} & \bf{82.11} & \bf{82.31}\\
        \hline
        
    \end{tabular}}
    \label{tab:offensive}
\end{table}

\begin{table}[]
    \caption{Macro-averaged (P)recision, (R)ecall and F1 for hate speech classification. * represent results obtained with different learning rate (2e-5) instead of 8e-5. C: Character n-grams W: Words n-grams.}

    \centering
    \scalebox{0.9}{
    \begin{tabular}{|c|c|c|c|c|}
        \hline

        \bf{Classifier} & \bf{Acc\%} & \bf{P} & \bf{R} & \bf{F1} \\ \hline\hline
        SVM (C) & 92.52 & \bf{84.71} & 71.12 & 75.76\\
        \hline
        SVM (W) & 90.08 & 73.90 & 70.73 & 72.15\\
        \hline
        SVM (C+W) & 92.37 & 82.94 & 72.17 & 76.16\\
        \hline\hline
        mBERT & 91.26 & 77.55 & 73.34 & 75.20 \\
        \hline
        XLM-RoBERTa$^*$ & 92.29 & 79.96 & 78.79 & 79.36\\ 
        \hline
        \hline
        
        AraBERT &  \bf{92.64} & 81.04 & \bf{79.31} & \bf{80.14}\\
        \hline
        QARiB & 92.99	& 82.99	& 77.72 & 80.04\\
        \hline
    \end{tabular}}
    \vspace{-0.2cm}
    \label{tab:hate}
\end{table}

Comparing the monolingual models, for offensive classification, we observe that mixed trained model -- QARiB, outperforms AraBERT by a margin of 2.3\%. Contrary to that, in hate classification, we noticed AraBERT - trained with formal text -- performs the best. However, the performance difference is very small ($0.1\%$) and can be attributed to the randomness mentioned in \citet{devlin-2019-bert}.\\

\paragraph{Error Analysis}
We analyzed classification errors of our best classifier for offensive language detection. Common cases are listed in Table \ref{tab:error-analysis-OFF}. We found similar errors in hate speech detection, and the main difference is that the target of the offense is a group of people as opposed to individuals. \\

\begin{table}[!htb]
    \caption{Common error types in offense detection. Classes: FP (False Positives) and FN (False Negatives)}
    \small
    \centering
    
    \begin{tabular}{c|c|r}
        \hline
        \rowcolor{LightGrey}
        \multicolumn{1}{c|}{\bf{CL}} & \multicolumn{1}{c|}{\bf{Error Type}} &  \multicolumn{1}{c|}{\bf{Example}}\\\hline
        
        \textbf{FP} & Annotation error & \<الله يفشلهن. ودك تتفل بوجهه>\\
        & & May God fail them. You wish to spit on his face\\

        \rowcolor{LightCyan}
         & Non-human target & \<هاشتاغ غبي.  وع مره اللون مقرف>\\
        \rowcolor{LightCyan}
        & & Stupid hashtag. Yucky! This color is very disgusting\\
        
        & Unclear context & \<احييك عالفكرة بس اذا كان وجهك>\\
        & & I salute you for the idea, but if your face\\
        
        \rowcolor{LightCyan}
        & Non-targeted offense & \<لا بارك الله للعدو. اللهم لا تؤاخذنا بما فعل السفهاء منا>\\
        \rowcolor{LightCyan}
        & (e.g., proverbs/idioms) & Oh God, do not bless the enemy (prayer)\\

        & Animal not offensive & \<اتوقع رسم قصة العنز وعياله. الكلاب كلوا القطوة>\\
        & & Expect to draw the story of the goat.. Dogs ate the cat\\

        \rowcolor{LightCyan}
        & Sarcasm & \<انا لكم الخت والام  اي حركه ا اذبحكم ههه>\\
        \rowcolor{LightCyan}
        &  & I'm your sister and mother. I will kill you haha\\\hline

        \textbf{FN} & Culture/Background & \<هذا مقدارك. اليوم زي وجهك>\\
        & & ``It's your destiny''. Today looks like your face (ugly)\\
        
        \rowcolor{LightCyan}
        & OOV/Unseen words & \<اتكتمي يابت منك ليها. خرا فيج زين>\\
        \rowcolor{LightCyan}\
        & & Shut up you and she. Shit on you!\\
        
        & Need understanding & \<الكدب ليس سياسة تحريرية ولكن إسلوب حياة>\\
        & & Lying is not editorial policy but a way of life\\
        
        \rowcolor{LightCyan}
        & Implicit insult & \<حضرة الدكتور فلاح. شوف مين الي بتتكلم>\\
        \rowcolor{LightCyan}
        & & The doctor is a villager (uncivilized). Look who talks! \\
        
        & Informal writings & \<ما كف بعطيككك بوكسس. إنت مرت زقة>\\
        & & I will givve youu a \textbf{blowww}. You're a \textbf{merce nary}\\
        
        \rowcolor{LightCyan}
        & Offensive emoji only& \<مش عاوز كلام كتير بس طرش له ده (الإصبع الوسطى) وبيسكت>\\
        \rowcolor{LightCyan}
        & & Just send (middle finger) to him and he will be silent\\
        
        & Negation & \<معد فيه حياء في البنت. قلة أدب>\\
        && The girl no longer has shyness. Incivility \\
\hline
        \end{tabular}

    \label{tab:error-analysis-OFF}
\end{table}

From the Table \ref{tab:error-analysis-OFF}, we noticed most of the confusion occurs due to lack of context in the input, implicit offensive instances, misunderstanding of grammatical construct like negation, bias towards some animals and the presence of sarcasm. Moreover, we also noticed some classification errors due to human annotation errors -- which is common given the complexity of the annotation task itself and its dependence of individual perspective.

\paragraph{Model Prediction Explanability}
We use Local Interpretable Model-Agnostic Explanations (LIME) (\cite{ribeiro2016why}) to interpret our best classifier. LIME is a model-agnostic method that perturbs the input text and gives weight to words based on how the model's predictions change.\\

\begin{figure}
\begin{center}
\includegraphics[scale=0.3]{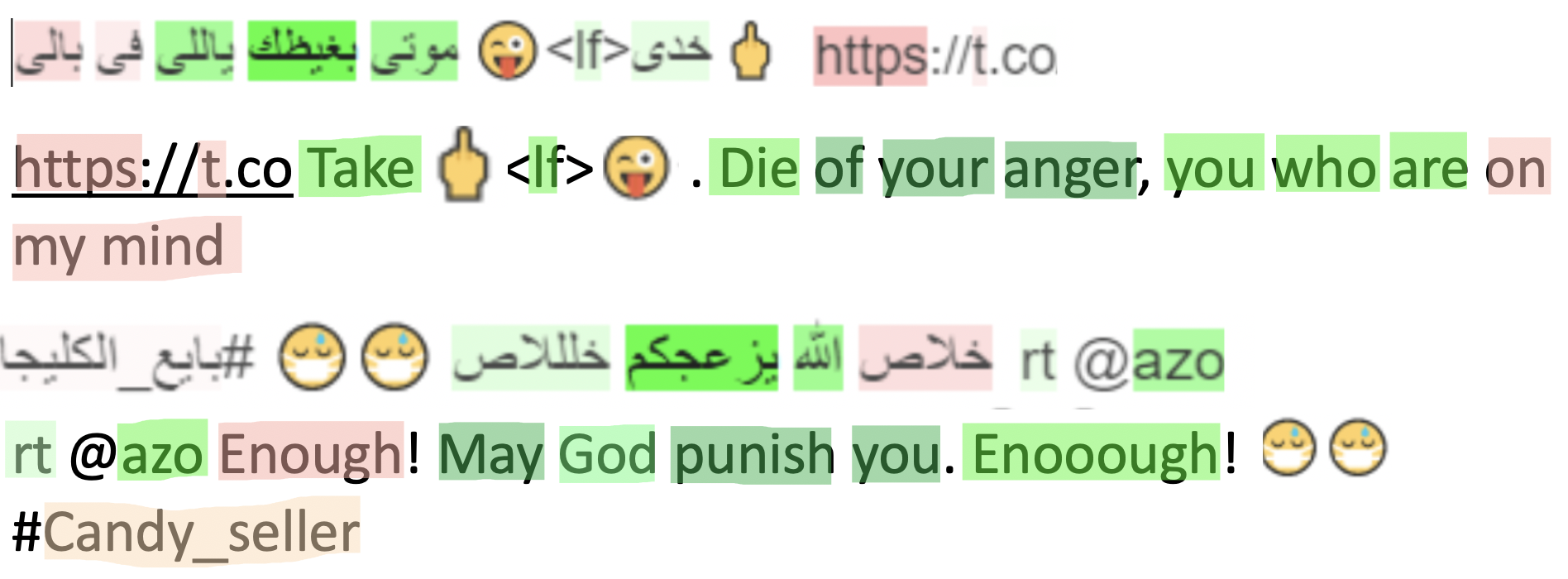} 
\vspace{1.1cm}
\caption{Examples of explainability errors. Green colored words contribute to classification as offensive while red colored words contribute to classification as non-offensive class. Rough translations are provided.}
\label{fig:lime}
\end{center}
\end{figure}

We analyzed a sample of
false-positive and 
false-negative examples from output of LIME and our best classifier. We found many confusion in model decision while considering the words as offensive or clean. We summarize common errors in the examples shown in Figure \ref{fig:lime} and recommend the followings for better explainability results: \textit{(i)} Ignore Twitter-specific symbols and symbols used for formatting (e.g. RT (Retweet), user mentions, links, <LF> (newline)), e.g., replace mentions with @USER, links with URL, <LF> with space.
    \textit{(ii)} As emojis are not recognized by LIME, replace them with their meanings, e.g., middle\_finger.
    \textit{(iii)} Remove repeated letters that are commonly found in tweets and used for emphasis, e.g., convert \<خلللاص> (``xlllAS'' - ``enooough'') to \<خلاص> (``xlAS'' - ``enough'').
    \textit{(iv)} We noticed confusion in interpreting words' contribution towards classifying (non-)offensiveness. However, such confusion is understandable given a limited resource that is used to learn discriminating linguistic information. In the  future, we plan to explore use of more balanced large data, collected from diverse sources.
    It's worth mentioning that those errors are not necessarily LIME errors as we have errors due to model predictions (F1 is 82.31).

    
    
    

\subsection{Generalization Capability of Models}
\label{ssec:generalization}

Our proposed method of collecting an offensive language dataset relies on emojis. While this method yields higher percentage of offensive content compared to existing methods, it's crucial that models trained on our dataset perform well not only on our test set but also on external datasets. Achieving such generalization can help us to assert with confidence that our method captures characteristics of offensive language that are universal, not specific to our dataset only.

In addition, assessing the generalization of the model to other social media platforms such as Facebook and YouTube, can indicate the universal usage of emoji and also effectiveness of harnessing the extralinguistic information present in emojis for offensive task.

\subsubsection{Generalization to Other Twitter Data}
To ascertain the generalization capability to other Twitter data, we evaluate our best performing model (QARiB, fine-tuned on our offensive training data) on the official test set of SemEval 2020 task 12 (OffensEval 2020, Arabic data) \cite{zampieri-etal-2020-semeval} data. For comparison, we also fine-tune QARiB on SemEval training data and evaluate on our dataset. Further, we fine-tune QARiB on combination of the two datasets and evaluate on the test sets separately. 

The SemEval data contains 8,000 tweets for training and 2,000 tweets for testing. The tweets are manually annotated for offensiveness. To have higher percentage of offensive tweets, only tweets that contained "yA" in text were considered. "yA" is commonly used in Arabic offensive language (\cite{zampieri-etal-2020-semeval}). 20\% of the data are tagged as offensive. \\

From Table \ref{tab:generalization}, we can see that QARiB, fine-tuned on our data, achieves F1 score of 85.21 on the SemEval test set. However, when fine-tuned on SemEval data but tested on our data, QARiB achieves F1 score of 72.26. This suggests that models trained on our dataset capture characteristics of SemEval data reasonably well while also having more variety compared to SemEval data. It's worth mentioning that combining the two datasets yield F1 score of 91.57 on SemEval dataset, which is an increase of 1.40  from the highest ranked system (\cite{alami-etal-2020-lisac}) at SemEval that achieved F1 score of 90.17 on the same test set. 

\begin{table}[]
    \caption{Performance comparison of QARiB, fine-tuned on our dataset (EMOJI-OFF), SemEval dataset and combination of the two datasets}
    \centering
    \scalebox{0.85}{
    \begin{tabular}{|c|c|c|c|c|c|}
        \hline
        \bf{Train Data} & \bf{Test Data} & \bf{Acc\%} & \bf{P} & \bf{R} & \bf{F1} \\ \hline \hline
        EMOJI-OFF & EMOJI-OFF & 84.02 & 82.53 & 82.11 & 82.31\\ \hline
        EMOJI-OFF & SemEval & 89.30 & 82.40 & 90.05 & 85.21\\ \hline
        \hline
        SemEval & EMOJI-OFF & 78.55 & 82.48 & 70.61 & 72.26\\ \hline
        SemEval & SemEval & 94.65 & 92.61 & 90.42 & 91.45\\ \hline
        \hline
        EMOJI-OFF + SemEval & EMOJI-OFF & 83.55 & 81.88 & 81.98 & 81.93\\ \hline
        EMOJI-OFF + SemEval & SemEval & 94.55 & 91.31 & 91.84 & 91.57\\ \hline

        \end{tabular}
        }

    \label{tab:generalization}
\end{table}

\subsubsection{Generalization to Multi-Platform Data}
Having established generalization capability of our model to external Twitter data, we are interested in assessing how much our model can generalize to other social media platforms such as Facebook and YouTube. This is a challenging task as content on these platforms can differ greatly from each other. For example, Twitter imposes a tweet length limit of 280 characters. There is no such limitation on Facebook or YouTube comments. The audience is also different on the different platforms. Because of these differences, user write in different style across the different platforms. Moreover, as \cite{chowdhury2020multi} point out, the percentage of offensive content is much higher on Twitter compared to YouTube or Facebook. 

\begin{table}[!htb]
    \caption{Performance on multi-platform data. MPOLD-TW/YT/FB refer to Twitter, YouTube and Facebook portions of MPOLD dataset respectively. The MPOLD-TW/YT/FB columns contain numbers reported in \cite{chowdhury2020multi}. The EMOJI-OFF column represents QARIB fine-tuned on our data. All numbers are macro averaged F1.}
    \centering
    \scalebox{0.8}{
    \begin{tabular}{|c||c||c|c|c|}
        \hline
        \bf{Test Data} & \bf{EMOJI-OFF} & \bf{MPOLD-TW} & \bf{MPOLD-YT} & \bf{MPOLD-FB}\\ \hline
        MPOLD-TW & 72.62 & - & 54 & 51\\ \hline
        MPOLD-YT & 68.2 & 60 & - & 53\\ \hline
        MPOLD-FB & 66.95 & 62 & 62 & -\\ \hline        

        \end{tabular}
        }

    \label{tab:multplat}
\end{table}

To assess our model performance on other platform content, we evaluate the MPOLD dataset (\cite{chowdhury2020multi}) using our best model (EMOJI-OFF). The MPOLD dataset is a multi-platform dataset that contains 4,000 Arabic news comments from Twitter (1,624), YouTube (1,592), and Facebook (784). These comments are manually tagged for offensiveness. Out of the 4,000 comments, 675 (16.88\%) were tagged as offensive. 


We report our cross-domain model performance in Table \ref{tab:multplat}. For better comparison, we present the cross-domain results reported by \cite{chowdhury2020multi}.
From Table \ref{tab:multplat}, we can see that QARiB, fine-tuned on our data, achieves F1 score of 68.2 and 66.95 on YouTube and Facebook portions of MPOLD dataset respectively. While these numbers are lower than that we saw earlier in Table \ref{tab:generalization} (as expected), they are still considerably higher than the cross-domain numbers reported in \cite{chowdhury2020multi}, which are 0.60 and 0.62 for YouTube and Facebook testsets respectively. This suggests that models trained on our data have the potential of generalizing to content on other platforms. 

\subsubsection{Generalization to Datasets without Emoji}
In order to assess how well our model performs on datasets without emoji, first, we removed all emojis from our dataset and fine-tuned our best model, QARiB. We evaluated this model on our test data and SemEval data (both with emojis removed). We did not notice any significant difference in performance as the model obtained F1 score of 85.03 on SemEval test set and F1 score of 83.05 on our test set, both of which are comparable to the numbers reported in Table \ref{tab:generalization}. In a second experiment, we retained the emojis in our training set, but excluded all tweets from the SemEval test set that contain emojis. Our model achieved F1 score of 83.94 in this setting. This again suggests, models trained on our data are not reliant on emojis and can generalize to datasets without emojis.

\section{Ethics and Social Impact}
\label{ethics}
After showcasing the effectiveness of using emojis for data collection and design, here we discuss the social impacts and ethical concern surrounding the newly released data.

\subsection*{User Privacy}
The dataset was collected using the Twitter API. For privacy, we anonymised the dataset by removing usernames and other sensitive user identifier.
However, in the future, if any concern is raised for a particular content, we will comply to legitimate concerns by removing the affected tweets from the corpus. Furthermore, to keep privacy and datasets integrity, we are providing tweets by text not the corresponding tweet ids. 

\subsection*{Biases}
Any biases found in the dataset are unintentional, and we do not intend to do harm to any group or individual. The bias in our data, for example towards a particular group is unintentional and is a true representation of the Twitter space during the period of our collection and according to our generic proposed method. Emojis used as seeds to create the collection are extracted automatically from publicly available datasets without any bias in our selection. Collection was obtained over a span of 18 months to have a good diversity of tweets written by many authors.\\

As for the assigned annotation labels, we follow a well-defined schema and available information to perceive final labels for offensiveness, hate speech and vulgarity. The labels are not a reflection of our opinion. Test questions used for quality control were selected to cover different annotation classes proportionally to their distributions in the collected tweets. \\

Although more than 190 workers have participated in the annotation process (with at least 80\% success ratio for 200 test questions), this doesn't guarantee perfect quality of annotation. As noted in \cite{mubarak2014using}, Arabic Twitter is dominated by tweets from Gulf countries (Saudi Arabia, Kuwait, etc.) while Appen (previously CrowdFlower/Figure8) crowdsourcing platform has $\approx 30\%$ of annotators from Egypt (the most Arabic populated country) (\cite{mubarak2016demographic}). This may lead to misunderstanding (and incorrect labeling) of portion of tweets especially when they are written in dialects that are not understood by annotators. Moreover, almost three quarters of Arab annotators on Appen are males which may lead to hidden biases in data annotation.

\subsection*{Potential Misuse}  
We urge the research community to be aware that our dataset can be used to misuse as like any other social media data. If such misuse is noticed, human moderation is encouraged in order to ensure this does not occur.

\section{Conclusion}
\label{conclusion}
The automation of detecting offensiveness (including hate speech) are limited by the availability of large balanced manually annotated dataset. To overcome such lacking, we propose a generic language-, topic- and genre-independent approach to collect a large percentage of Arabic offensive and hate speech data.
We utilize the extralinguistic information embedded in the emojis to 
collect a large number of offensive tweets. 
We manually annotated the Arabic dataset for different layers of information, including \textit{offensiveness}, followed by fine-grained hate speech, vulgar and violence content annotation. 
Furthermore, we benchmark the dataset for detecting offensive content and hate speech using different transformer architectures and performed in-depth analysis for temporal changes, linguistic usage and target audience. We also studied specific patterns in violence tweets. We demonstrate that the transformer model trained on our data achieves strong results on an external Twitter dataset. Further, the model outperforms previous models in literature when tested on multi-platform data that contains offensive comments from Twitter, YouTube and Facebook.
This is the largest Arabic dataset for multilevel offensive language detection annotated with fine-grained hate speech, vulgar and violence labels. We publicly release the dataset for the research community.

\paragraph{Competing interests:} The author(s) declare none
\pagebreak

\bibliography{custom}
\bibliographystyle{nlelike}

\end{document}